%% file: fullPaper.tex
\documentclass{multibody2023_full_paper}

\usepackage{graphics} 
\usepackage{graphicx}
\usepackage{url}
\usepackage{amsmath}
\usepackage{amsfonts}
\usepackage{xcolor}
\usepackage{listings}
\usepackage{subcaption}
\usepackage{multirow}
\usepackage{siunitx}
\usepackage{placeins}

\colorlet{RED}{red}

\include{macros}

\begin{document}

\begin{center}
  \fontsize{14}{16}{\bf
  Using simulation to design an MPC policy for field navigation using GPS sensing
  }
\end{center}
 
\begin{center}
\normalsize{
  \bf{
    Harry Zhang$^{1}$,  
    \underline{Stefan Caldararu$^2$}, 
    Ishaan Mahajan$^2$,
    Shouvik Chatterjee$^1$,
    Thomas Hansen$^3$,
    Abhiraj Dashora$^2$, 
    Sriram Ashokkumar$^2$, 
    Luning Fang$^1$,
    Xiangru Xu$^1$,
    Shen He$^4$ 
    Dan Negrut$^1$}
}
\end{center} 


\begin{center}
  \begin{tabular}{c}
    $^1$ Department of Mechanical Engineering                        \\
    University of Wisconsin-Madison                                  \\
    Madison WI, USA                                                  \\
    {[hzhang699, schatterjee9,lfang9}\\
    {xiangru.xu, negrut]}@wisc.edu \\
  \end{tabular}
  \begin{tabular}{c}
    $^2$ Department of Computer Sciences                    \\
    University of Wisconsin-Madison                         \\
    Madison WI, USA                                         \\
    {[scaldararu, imahajan, }
    \\{dashora, ashokkumar2]}@wisc.edu \\	
  \end{tabular} \\ 
  \begin{tabular}{c}
    $^3$Department of Electrical \\and Computer Engineering \\
    University of Wisconsin-Madison                       \\
    Madison WI, USA                                       \\
    thansen8@wisc.edu                                     \\  
  \end{tabular}
  \begin{tabular}{c}
    $^4$Department of Mechanical Engineering \\
    California State University-Los Angeles\\
    Los Angeles, California, USA                                       \\
    he.shen@calstate.edu                                    \\  
  \end{tabular}
\end{center}


\begin{quote} 
	
\section*{ABSTRACT}
Modeling a robust control system with a precise GPS-based state estimation capability in simulation can be useful in field navigation applications
as it allows for testing and validation in a controlled environment. This testing process would enable
navigation systems to be developed and optimized in simulation with direct transferability to real-world scenarios.
The multi-physics simulation engine \CHRONO\, allows for the creation of scenarios that may be difficult or dangerous to 
replicate in the field, such as extreme weather or terrain conditions.
Autonomy Research Testbed (ART), a specialized robotics algorithm testbed, is operated in conjunction with \CHRONO\, to develop an MPC control policy as well as 
an EKF state estimator. This platform enables users to easily integrate
custom algorithms in the autonomy stack. This model is initially developed and used in simulation
and then tested on a twin vehicle model in reality, to demonstrate the transferability between simulation and reality (also known as Sim2Real).\\
\textbf{Keywords:} Sim2Real, Robotics, State Estimation, Model Predictive Control, Vehicle Dynamics, GPS Noise Model.

\end{quote}

\section{INTRODUCTION}
\label{sec:intro}
\input{sections/intro.tex}

\section{METHODS}
\label{sec:methods}
\input{sections/methods.tex}

\section{EXPERIMENTS}
\label{sec:experiments}
\input{sections/exp.tex}

\section{ANALYSIS}
\label{sec:analysis}
\input{sections/analysis.tex}

\section{CONCLUSION AND FUTURE WORK}
\label{sec:conclusion}
\input{sections/conclusion.tex}

\section*{ACKNOWLEDGMENTS}
This work was carried out in part with support from National Science Foundation projects CPS1739869, CISE1835674, and OAC2209791.

\bibliographystyle{splncs} 	

\bibliography{BibFiles/refsSensors,BibFiles/refsMachineLearning,BibFiles/refsAutonomousVehicles,BibFiles/refsChronoSpecific,BibFiles/refsSBELspecific,BibFiles/refsMBS,BibFiles/refsCompSci,BibFiles/refsTerramech,BibFiles/refsFSI,BibFiles/refsRobotics,BibFiles/refsDEM}
\end{document}

%% file: macros.tex
\newcommand{\softpackage}[1]{{\sffamily{#1}}}
\newcommand{\CHRONO}{{\softpackage{{Chrono}}}}
\newcommand{\CHRONOVEH}{{\softpackage{{Chrono::Vehicle}}}}
\newcommand{\CHRONOSEN}{{\softpackage{{Chrono::Sensor}}}}

\newcommand{\vect}[1]{\mathbf{#1}}
\newcommand{\matr}[1]{\mathbf{#1}}

%% file: sections/intro.tex
This paper reports results generated with a research testbed whose purpose is the characterization, measurement, and mitigation of the gap between simulation and reality. The simulation-to-reality gap, also known as the sim2real gap~\cite{sim2realGapEssex1995}, often prevents a robot's autonomy stack from having results in reality as expected from simulation. The potential causes for sim2real gap include: difference between simulation and reality dynamics; different sensor behaviors between simulation and reality; and different compute powers available in simulation and reality~\cite{hofer2021sim2real}. To close the simulation-to-reality gap, the simulation should consider and model the adverse conditions encountered in the real world applications. To against this backdrop, we propose a novel way of modeling GPS noise that better represents the noise in reality, in comparison to a normal distribution noise model typically used. We demonstrate the capabilities of a basic Extended Kalman Filter (EKF) working with a Model Predictive Control unit (MPC) tasked with tracking waypoint-based trajectories in simulation. Finally, we demonstrate the direct transferability by running the same autonomy stack on the real world twin of our vehicle, tracking it with a Motion Capture (MOCAP) system.\\
\begin{figure}[h]
\begin{center}
    \includegraphics[width=9cm]{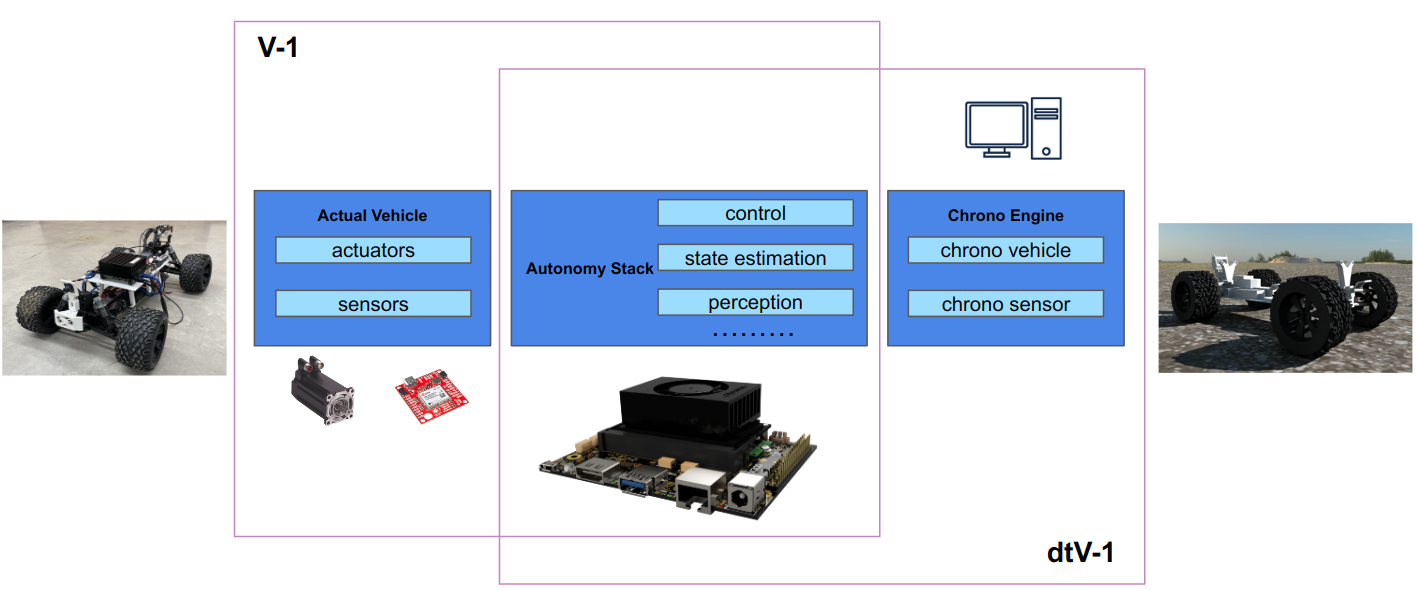}
\end{center}
\caption{Autonomy Research Testbed}
\label{fig:ART_dART}
\end{figure}

Autonomy Research Testbed (ART) is a platform providing a structure for autonomous vehicle algorithm development, as shown in Fig.~\ref{fig:ART_dART}. The real autonomous vehicle (V-1) as well as its digital twin (dtV-1) are used for the development of our control and state estimation algorithms~\cite{artatk2022}. V-1 is a 1:6 scaled vehicle, equipped with a NVIDIA Jetson AGX, and a variety of embedded systems and sensors mounted. For the simulation, we use a high-fidelity multi-physics simulation engine, \CHRONO~\cite{chronoOverview2016}. \CHRONOVEH~\cite{chronoVehicle2019} and \CHRONOSEN~\cite{asherSensors2020} are leveraged to generate a virtual vehicle in simulation.

The same autonomy stack and the same computing equipment (NVIDIA Jetson AGX) are applied to both the simulated and real vehicle to directly analyze performance differences regarding vehicle dynamics and sensor behavior, and to mitigate the difference. The ART platform allows for autonomous algorithm development on the simulated vehicle with direct application to the real one, lowering the experimental cost. 

The work proposed here does not focus on novel state estimation algorithms nor control policies, but rather demonstrates the possibility of using the ART platform to design and test an autonomy stack in simulation, and then have direct application in reality with minimal testing. The paper is organized as follows: Sec.~\ref{sec:methods} presents EKF and MPC formulation as well as a new GPS noise model; Sec.~\ref{sec:Sim} talks about simulation for EKF with constant control inputs and simulation tests for MPC using a ``ground truth'' sensor providing accurate states, and then combining EKF and MPC to run virtual GPS-based navigation in simulation; Sec.~\ref{sec:mocap_testing} does the sim2real comparison; finally Sec.~\ref{sec:analysis} and \ref{sec:conclusion} are our analysis, conclusion and future plans for this work.

%% file: sections/methods.tex
\subsection{Vehicle dynamics model}
\label{sec:motionModel}
A 4-DOF basic bicycle dynamics model is derived for both the MPC formulation and the EKF. The state variable, $\mathbf q = [x,y,\theta,v]^T$, includes the Cartesian coordinates $x$ and $y$, heading angle $\theta$, and the speed of the vehicle $v$. The model updates via a control input $\mathbf u = [\alpha,\delta]^T$, consisting of the vehicle's throttle and steering angle input, respectively. The time derivative of the state variable $ \dot{\mathbf q}$ is approximated as follows: 

\begin{equation}
    \label{equ:dynamics}
    \dot {\mathbf q} = f(\mathbf q,\mathbf u) = 
    \begin{pmatrix}
        v \cos(\theta) \\
        v \sin(\theta) \\
        \frac{v \tan(\delta)}{l} \\
        \frac{R_{wheel} \gamma}{I_{wheel}} [f_1(v, \alpha) - f_0(v)]
    \end{pmatrix}
\end{equation}
where $R_{wheel}$ and $I_{wheel}$ are the radius and inertia of the wheel respectively, $l$ is the wheelbase and $\gamma$ is the gear ratio. The time derivative of the vehicle speed $\dot{v}$ is related to the torque generated from the DC motor, which is represented by $f_1(v,\alpha)$, as well as the motor resistance torque, $f_0(v)$. Here, we define $f_1(v,\alpha)$ and $f_0(v)$ as:
\begin{subequations}
	\begin{align}
	f_1(v,\alpha) &= \tau_0\alpha - \frac{\tau_0 v}{\omega_0 R_{wheel} \gamma}, \label{equ:motor_dynamics} \\
	f_0(v) &= \frac{vc_1}{R_{wheel} \gamma}+ c_0. \label{equ:motor_resistance}
	\end{align}
\end{subequations}
Herein, $\tau_0$ is the stall torque, $\omega_0$ is the maximum angular velocity of the motor, and $c_0$ and $c_1$ are the coefficients for approximating motor resistance. The simplified 4-DOF model allows fast optimization in the MPC formulation, and is also used as the state-update for the EKF formulation.

\subsection{EKF Formulation}
\label{sec:EKF}
We adopted the Extended Kalman Filter proposed in \cite{elementaryKalmanFiltering2017}, which follows a 2 stage approach to estimate the state of the vehicle. The first stage involves prediction of the next state given the previous state and the current control input, as shown in Eq.(\ref{equ:stateupdate}).
\begin{equation}
	\label{equ:stateupdate}
	\mathbf q_{t+1|t} = \mathbf q_{t|t}+\mathbf f(\mathbf q_{t|t},\mathbf u).
\end{equation}
The predicted covariance estimate is determined using the linearized version of the state update equation and an error matrix. The Jacobian of the motion model in Eq.(\ref{equ:dynamics}) is evaluated as follows:
\begin{equation}
    \label{equ:motModJacob}
    \mathbf F = 
    \begin{pmatrix}
        1 & 0 & -v \sin \theta & \cos \theta \\
        0 & 1 & v \cos \theta & \sin \theta \\
        0 & 0 & 1 & \frac{\tan \delta}{l} \\
        0 & 0 & 0 & 1-\frac{(\tau_0/\omega_0)+c_1}{I_{wheel}}
    \end{pmatrix}.
\end{equation}

The second stage involves correction of the predicted position by an observation model with sensor inputs. Given the motion model described above, this follows the standard EKF formulation.
The filter allows for specification of a starting position and heading, and generates a Local Tangent Plane (LTP) correspondingly.
All future measurements are provided relative to the LTP.
After correction, the current state estimate is sent to the MPC controller and the filter restarts the cycle once it receives a new control input and new measurements.

\subsection{MPC Formulation}
\label{sec:MPC}
\begin{figure}[h]
    \begin{center}
        \includegraphics[width=6cm]{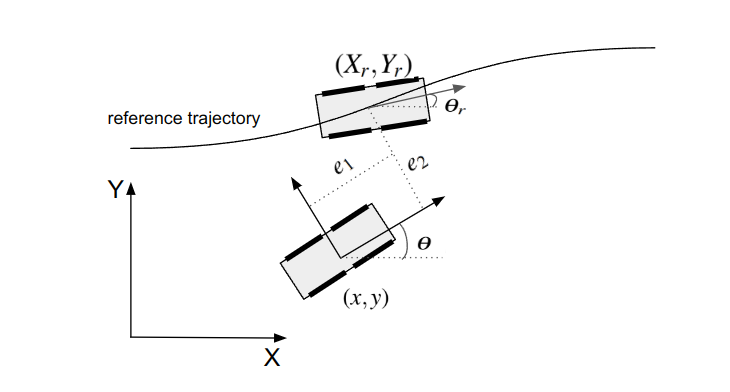}
    \end{center}
    \caption{Error dynamics of a vehicle following a certain trajectory.}
    \label{fig:err_dynamics}
\end{figure}

The MPC formulation is based on solving a trajectory tracking problem using the error dynamics~\cite{klanvcar2007tracking}. As shown in Fig.~\ref{fig:err_dynamics}, an error state, $\bf e$, is defined as follows:
\begin{equation}
\label{equ:error_state}
\bf e=
\begin{pmatrix}
e_1\\
e_2\\
e_3\\
e_4
\end{pmatrix}
=
\begin{pmatrix}
\cos\theta & \sin\theta & 0 & 0\\
-\sin\theta & \cos\theta & 0 & 0\\
0 & 0 & 1 & 0\\
0 & 0 & 0 & 1
\end{pmatrix}
\begin{pmatrix}
x_r-x\\
y_r-y\\
\theta_r-\theta\\
v_r-v
\end{pmatrix}
\end{equation}
where $\mathbf{q}_r = [x_r, y_r, \theta_r, v_r]^T $ is the predefined ideal reference. As the vehicle state $\mathbf q$ is updated at each time step, a corresponding reference state $\mathbf q_r$ will be determined as well.

Here, trajectory tracking based MPC can be set up by solving an optimal control problem over finite prediction horizon~\cite{rokonuzzamanMPCBicycleModel2021}. More details on the derivation based on the vehicle model and trajectory tracking problem can be found in \cite{TR-2023-01}. The optimal control problem is formulated as follows:
\begin{subequations}
    \begin{equation*}
    J_t^*(\vect{e}_t) = \min_{ \mathbf u_k} \;\;\mathbf e_N^T \mathbf Q \mathbf e_N + \sum_{k=0}^{N-1} \mathbf e_k^T \mathbf Q \mathbf e_k + (\mathbf u_k- \mathbf u_r)^T \mathbf R (\mathbf u_k-\mathbf u_r)
    \end{equation*}
    \begin{equation}
        \vect{e}_{k+1} = \matr{A}_k \;  \vect{e}_{k} + \matr{B}_k \; \vect u_k 
    \end{equation}
    \begin{equation*}
        \mathbf A_k = 
        \begin{pmatrix}
           0 & \frac{v_k \; \tan{\delta_k}}{l} & -v_r \; \sin{e_3} & 0 \\
           -\frac{v_k \; \tan{\delta_k}}{l} & 0 & v_r \; \cos{e_3} & 0 \\
           0 & 0 & 0 & 0 \\
           0 & 0 & 0 & -\frac{c_1 \omega_0 + \tau_0}{I_{wheel} \omega_0}  
        \end{pmatrix} \; \Delta t + \mathbb{I}_{4 \times 4}\;\;\;\;\;\;
        \mathbf B_k = 
        \begin{pmatrix}
            0 & \frac{v_k \; e_2}{l \; \cos^2{\delta_k}}\\
            0 & -\frac{v_k \; e_1}{l \; \cos^2{\delta_k}} \\
            0 & -\frac{v_k}{l \; \cos^2{\delta_k}}\\
            -\frac{\tau_0 R_{wheel} \gamma}{I_{wheel}} & 0
        \end{pmatrix}
        \; \Delta t
        \end{equation*}
    \begin{equation}
        \vect{e}_{k} \in \matr{E}, \mathbf u_{k} \in \mathbf U, k=0,..,N-1
    \end{equation}

\end{subequations}
Here, $\mathbf Q \in \mathbb{R}^{4 \times 4}$and $\mathbf R\in \mathbb{R}^{2 \times 2}$ are the weight matrix for different components in the optimal error state and control inputs; $N$ is the prediction horizon; $\mathbf E$ and $\mathbf U$ are the ranges of error states and control inputs. At every time step, an optimal control problem is solved using the OSQP package~\cite{stellatoOSQP2020}.


\subsection{Simulated noise model}
\label{sec:noiseModel}
\begin{figure}[h]
   \centering
   \subfloat[\centering Real Measurements \label{fig:RGPS}]{{\includegraphics[width=4cm]{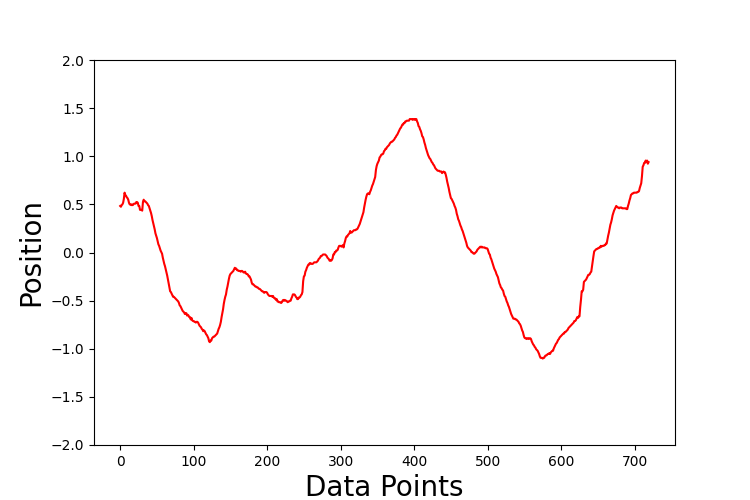} }}%
   \qquad
   \subfloat[\centering Noise Model Measurements \label{fig:SGPS}]{{\includegraphics[width=4cm]{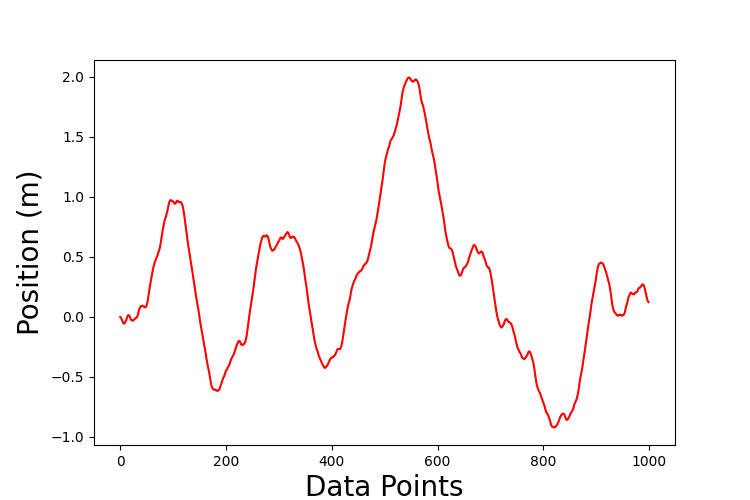} }}%
   \caption{GPS data in Meters for a single point, in one dimension}%
   \label{fig:GPSModel}
\end{figure}

A novel noise model described in~\cite{GPSNoiseModel2022} is used to simulate GPS data more accurately. Therein, the model uses 
a random walk in the second derivative of the position, with a concentration gradient on the normal distribution. By allowing for the normal distribution to be in the second derivative, the model accurately reflects the smooth transitions
of GPS measurements, and the concentration gradient ensures an expectation of 0-mean. The equations describing this model are provided in Eq.(\ref{equ:noiseModel}). Here, we have that $p_{t+1}$ is the noise that will be added to the measurement. $a_{t+1}$, the second derivative of our noise model is just a normal distribution with $m_{t+1}$ as the mean, and some constant $\sigma$ as the standard deviation. $m_{t+1}$ depends on the maximum noise value that is desired as well as the current noise level, ensuring an expectation of 0-mean. Finally, $v_{t+1}$ is just a variable that maintains the first derivative of our noise distribution. A comparison of simulated and real data is provided in Fig.~\ref{fig:GPSModel}.
Figure~\ref{fig:RGPS} provides real GPS measurements, at one stationary point. The GPS measurements are converted to Cartesian coordinates via an LTP centered at the initial measurement, and only the X coordinates are displayed. 
Figure~\ref{fig:SGPS} shows the simulated noise model, with an initial noise measurement of \SI{0}{\meter}, and with the maximum noise set to around \SI{2}{\meter}. As can be seen in Fig~\ref{fig:RGPS} new measurements do have dependency on old measurements, which is more accurately reflected by our noise model as opposed to a standard normal distribution noise model.

\begin{equation}
	\begin{split}
		m_{t+1} = \frac{-p_t}{p_{max}}\\
		a_{t+1} = N(m_{t+1},\sigma)\\
		v_{t+1} = v_t+a_{t+1}\\
		p_{t+1} = p_t+v_t+a_{t+1}
	\end{split}
	\label{equ:noiseModel}
\end{equation}

\FloatBarrier
The reason for proposing this new GPS noise model is because the ``priviliged information'' provided to the MPC controller in simulation does not exist in reality. To further gauge the collective performance of the autonomy stack, the MPC algorithm should be tested utilizing the filtered noisy data from the EKF. In most of the GPS navigation situations, GPS sensors will normally produce a noise with a magnitude of 1-2 \SI{}{\meter}. Without capturing the noise from the GPS sensor, the simulated results of the control policies can not be reliably reproduced in reality, as shown in~\cite{Shi2023Marl, quinlan2010SimToLife}. To better capture the randomness of GPS drifting noise and facilitate simulation fidelity, the noise model we propose above will produce data provided to the Extended Kalman Filter in order to execute state estimation in simulation. 

%% file: sections/exp.tex
Rather than migrating directly from simulation to an outdoor environment, we take an intermediate step by conducting experiments using a motion capture system (MOCAP)~\cite{furtado2019comparative} indoor. The MOCAP system provides high accuracy position information within a confined indoor space, allowing for accurate quantification of the EKF and MPC performance.

In Section~\ref{sec:Sim}, the autonomy stack is developed fully in simulation. Next, for each experiment, the same autonomy stack developed in simulation is deployed on the real vehicle. Moreover, the same embedded system, an NVIDIA Jetson AGX, was used to run the autonomy stack. Specifically, in reality, the Jetson card was attached to the physical vehicle; while in simulation, the Jetson was ``attached'' to a workstation that runs vehicle and sensor simulation in \CHRONO\, and the rest of the virtual world. Multiple experiments were conducted for each scenario to demonstrate the accuracy of simulation.
\subsection{Simulation}
\label{sec:Sim}

\subsubsection*{EKF}
\label{sec:EKFsim}
\begin{figure}[h]

    \begin{subfigure}[b]{0.29\textwidth}
    \centering
    \includegraphics[width=\textwidth]{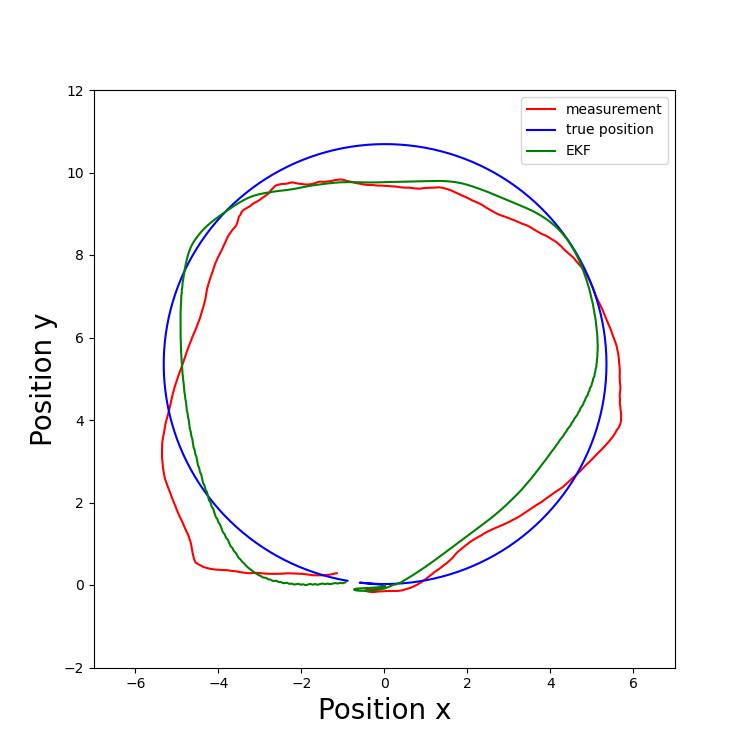} %
    \caption{EKF with circle trajectory}
    \label{fig:sim_ekf}
    \end{subfigure}
    \hfill
    \begin{subfigure}[b]{0.29\textwidth}
    \centering
    \includegraphics[width=\textwidth]{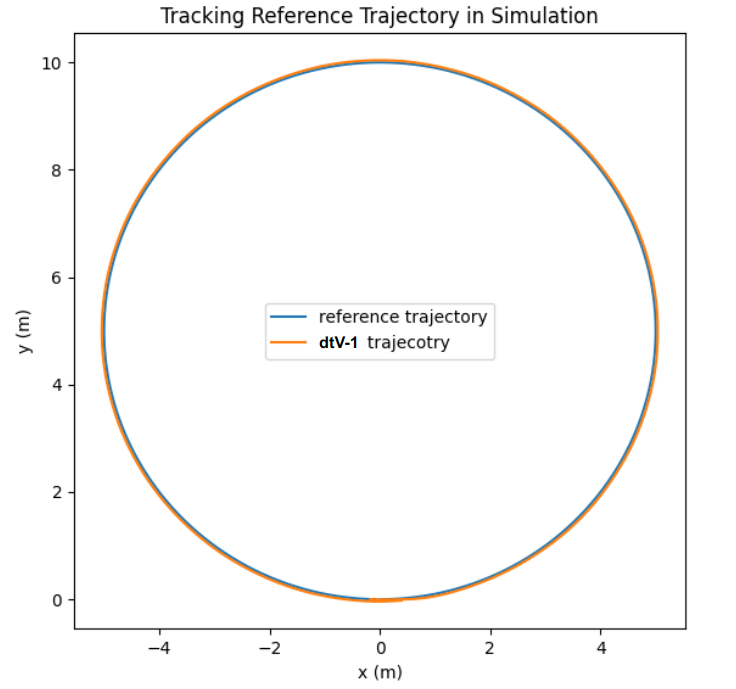} %
    \caption{MPC tracking circle trajectory}
    \label{fig:sim_mpc_circle}
    \end{subfigure}
    \hfill
    \begin{subfigure}[b]{0.35\textwidth}
    \centering
    \includegraphics[width=\textwidth]{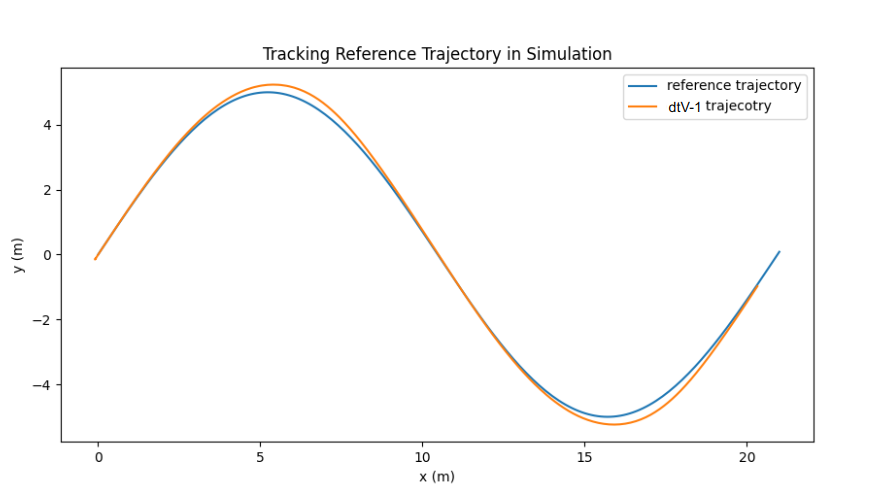} %
    \caption{MPC tracking sinusoidal trajectory}
    \label{fig:sim_mpc_sin}
    \end{subfigure}
    \caption{Simulation results for EKF and MPC}
    \label{fig:sim_ekfandmpc}
\end{figure}

Here, we discuss the simulation of our EKF model. We provide only one test case as when the EKF is running independently, there are a small number of potential tests. By having a control sequence that does not depend on the state generated by the EKF, we are able to measure the error generated exclusively by the EKF. A single circle test is demonstrated in Fig.~\ref{fig:sim_ekf}.\
We found that while the simplified motion model described in section~\ref{sec:motionModel} is able to accurately capture the dynamics of the vehicle with constant control inputs, it struggles to do so for control inputs with high variability (as given by a true MPC). As a result the EKF is tuned in simulation for performance with the MPC control inputs placing higher weight on the measurement and may not have optimal performance on the circle. 

In Table~\ref{tab:ekf_data}, we show data for ten tests run on the above circle. For each test, the average distance between the measurement and ground truth, as well as EKF prediction and ground truth is shown. In addition, the maximum deviation is also displayed. For nine of the ten tests, the EKF had better maximum error, and for eight it had better average error.

\begin{table}[!htb]
    \begin{minipage}{.5\linewidth}
      \centering
      \begin{tabular}{|c|c|c|c|c|}
        \hline
        & \multicolumn{2}{c|}{maximum error} & \multicolumn{2}{c|}{average error} \\
        \hline
        tests: & EKF & MEAS & EKF & MEAS\\
        \hline
        1 & 1.509 & 1.504 & 0.819 & 0.791 \\
        \hline
        2 & 3.553 & 3.935 & 1.544 & 1.579 \\
        \hline
        3 & 1.803 & 1.878 & 1.024 & 0.856 \\
        \hline
        4 & 1.101 & 1.379 & 0.609 & 0.772 \\
        \hline
        5 & 1.560 & 2.033 & 0.849 & 1.174\\
        \hline
        6 & 1.431 & 2.066 & 0.823 & 0.978 \\
        \hline 
        7 & 1.992 & 2.235 & 0.969 & 1.288\\
        \hline
        8 & 1.642 & 2.047 & 0.767 & 0.886\\
        \hline
        9 & 2.168 & 2.993 & 1.058 & 1.274\\
        \hline
        10 & 2.015 & 2.338 & 0.890 & 1.109\\
        \hline
    \end{tabular}
        \caption{EKF and measurement errors (m)}
        \label{tab:ekf_data}
    \end{minipage}%
    \begin{minipage}{.5\linewidth}
        \centering
        \begin{tabular}{|c|c|c|c|c|c|}
          \hline
          && \multicolumn{2}{c|}{maximum error} & \multicolumn{2}{c|}{average error} \\
          \hline
          trajectory: &tests: & SIM & REAL & SIM & REAL\\
          \hline
          1& 1 & 0.444 & 0.133 & 0.235 & 0.058\\
          \hline
          1& 2 & 0.385 & 0.452 & 0.165 & 0.236\\
          \hline
          1& 3 & 0.857 & 0.140 & 0.428 & 0.072 \\
          \hline
          1& 4 & 1.441 & 0.431 & 0.687 & 0.234 \\
          \hline
          1& 5 &0.476 & 0.430 & 0.258 & 0.224\\
          \hline
          2& 1 & 0.545 & 0.356 & 0.203 & 0.213 \\
          \hline
          2& 2 & 0.302 & 0.606 & 0.171 & 0.287 \\
          \hline
          2& 3 & 0.557 & 0.160 & 0.248 & 0.076 \\
          \hline
          2& 4 & 0.326 & 0.175 & 0.135 & 0.095 \\
          \hline
          2& 5 & 0.818 & 0.618 & 0.334 & 0.253\\
          \hline
      \end{tabular}
          \caption{Sim and Reality testing (m)}
          \label{tab:sim_data}
    \end{minipage} 
\end{table}

\subsubsection*{MPC}
\label{sec:MPCsim}
To evaluate the performance of the MPC controller on tracking reference trajectories, privileged information directly from the simulation is provided to the dtV-1, instead of using the EKF estimation. As shown in Fig.(\ref{fig:sim_mpc_circle}) and (\ref{fig:sim_mpc_sin}), the MPC controller behaves well for both the circular and sinusoidal trajectories with the reference speeds set to \SI{1}{\m/s}, demonstrating the robustness and accuracy of the waypoint-based MPC tracking controller. 

\subsubsection*{EKF + MPC}
\label{sec:ekf_mpc_testing}

The following simulation scenario, that combines both the EKF and noise model (as Sec.~\ref{sec:noiseModel}) along with the MPC tracking controller, has the best chance to accurately simulate reality. The random walk noise model has been implemented on a virtual GPS sensor in the \CHRONO ~simulation engine~\cite{asherSensorSimulation2021}. For this simulation, the EKF takes inputs from virtual GPS and Magnetometer sensors, as well the steering and throttle inputs for the dtV-1. Then the EKF passes the estimated vehicle states to the MPC tracking controller to perform the trajectory tracking tasks. Since it is in the simulation, it is easy to access the ``ground truth'' data that indicates the actual trajectory of the vehicle. 

\begin{figure}[h]
    \centering
    \subfloat[\centering Reality Testing 1 \label{fig:reality_sin1}]{{\includegraphics[width=7.0cm]{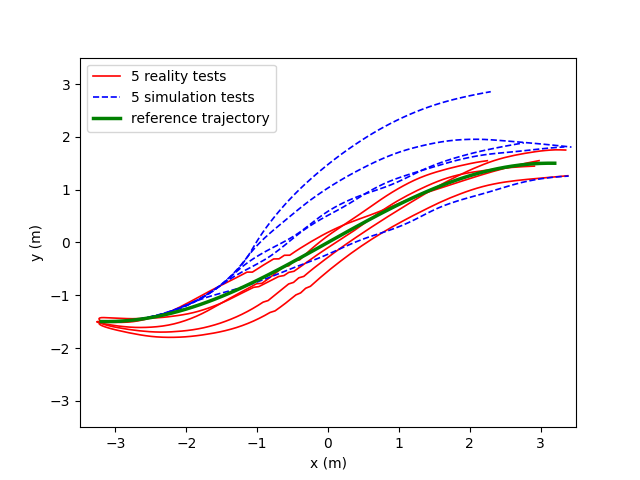} }}%
    \qquad
    \subfloat[\centering Reality Testing 2 \label{fig:reality_sin2}]{{\includegraphics[width=7.0cm]{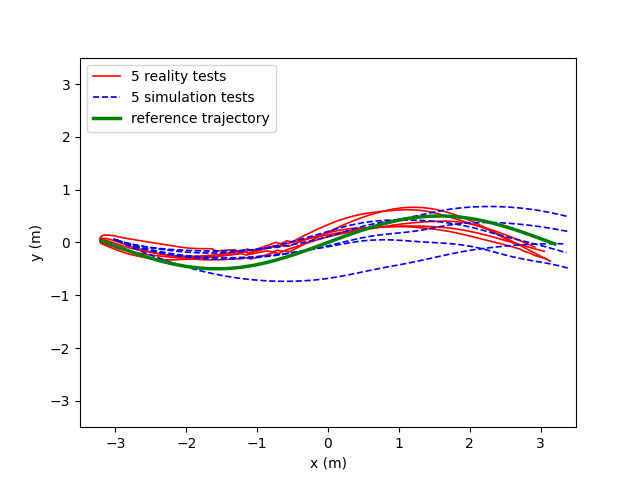} }}%
    \qquad
    \caption{Sim2Real result}%
    \label{fig:sim2real}
\end{figure}

In Fig.~\ref{fig:reality_sin1} and Fig.~\ref{fig:reality_sin2} we show two different trajectories, with ground truth results both in simulation and reality.

\label{sec:testingSim}

\FloatBarrier

\subsection{Reality}
\label{sec:mocap_testing}
Since the MOCAP provides precise position information, we are able to access and record trajectories (with error less than \SI{1}{\milli\meter}) that the vehicle goes through. However, to best reproduce the common GPS navigation scenario, the vehicle only accesses corrupted position information as described in~\ref{sec:noiseModel}. This noisy position data is made by superposing the ``Random Walk'' noise on the accurate MOCAP position information, for both the x and y coordinates. One drawback for conducting experiments with the MOCAP system is the limitations for vehicle's motion, which for our case is approximately $\SI{6}{\meter} \times \SI{3}{\meter}$. Correspondingly, the reference trajectories are shorter than the ones desired in outdoor environment. Figure~\ref{fig:sim2real} plots the real vehicle trajectories in red, simulated vehicle trajectories in blue, and the reference trajectory in green. Additonally, Table~\ref{tab:sim_data} shows the error in trajectory of the vehicle relative to the reference trajectory. This is the ground truth data, either provided by the MOCAP system or by the simulation. We show the error for all 5 tests for both trajectories in both simulation and reality. The maximum error is the largest deviation from the reference trajectory, and the average error is the average amoungst all recorded data points throughout the test. For each measurements, the shortest distance to the reference path is used.


%% file: sections/analysis.tex
According to previous studies~\cite{Shi2023Marl} on vehicles with GPS navigation, it is suggested that improvements on the Sim2Real gap may come in two forms. The first is improvement of the simulated vehicle's dynamics model. Additionally, previous experiments in simulation are conducted with no GPS noise, and compared to tests in reality with a standard GPS. We offer improvements on both fronts. The simulated vehicle dtV-1 is accurately calibrated to the real vehicle V-1, and additionally functions within the high fidelity simulation engine \CHRONO. Further, we simulate GPS noise using a new model, designed to more accurately replicate real GPS measurements.

As shown in Fig.~\ref{fig:sim2real}, the vehicle's MPC controller generally does a good job tracking the reference trajectory in both simulation and reality based on data in Table~\ref{tab:sim_data}. As can be easily observed, there are larger errors in trajectories in simulation than in reality. We noticed that this is likely due to a discrepancy in measurement noise between simulation and reality. In simulation, an average error of 0.73~m between the measurement and ground truth was observed, while in MOCAP testing there was only an error of 0.46~m between the measurement and reality, which likely caused by the difference in updating frequency in the noise models. While in simulation the noise model updates within the simulation regardless of the autonomy stack reading the data, the only place to corrupt the accurate MOCAP data is within the autonomy stack which has a much lower update frequency than the simulation. With a short trajectory, this does not allow for the noise model to deviate much as the measurement starts each test with a noise of 0. The results shown above are still positive, as the MPC is able to track the reference trajectory well in reality, and the EKF still provides higher accuracy than using just the raw measurement data.

While the MPC was able to perform well in both simulation and reality given the ``accurate'' data (in simulation testing with ground truth information, and in reality testing with relatively low noise), the EKF was unable to produce similar results. In over 7 of the 10 tests in the MOCAP lab, the EKF had worse average error than the raw measurements. However, the results match fairly well with simulation. When testing the EKF with smaller levels of noise in the measurement, it was noticed that the 4DOF motion model had limitations in its ability to predict the vehicles trajectory. When the noise levels in measurement were below a certain threshold, the motion model served only to add additional noise to the measurement. Another possible reason for this performance discrepancy is the Sim2Real gap, but analysis of the exact cause is difficult without a more advanced model than the simple 4DOF model.

%% file: sections/conclusion.tex
In this work, we utilize an Autonomy Research Testbed (ART) platform in conjunction with a high-fidelity simulation engine, \CHRONO\,, to develop and test autonomy algorithms in both simulation and reality. The positive aspects for this Sim2Real work include: utilizing high-fidelity vehicle simulation and novel GPS noise model. The framework demonstrates the potential of autonomy algorithm development in simulation by mitigating the Sim2Real gap~\cite{Shi2023Marl}.

Although we observed progress in the mitigation of the Sim2Real gap, it still exists in our experiments. While the purpose of this work is not to advance EKF or MPC design, improvements in the algorithms may offer better insights into the differences between simulation and reality. Although the 4DOF motion model functions well for the MPC as it uses a linearized model with a small time step, the EKF may benefit from a more robust model. This would allow for better characterization of errors produced by the Sim2Real gap, as opposed to those inherently produced by the model. In the future, we hope to improve the state estimation model through implementation of a more complicated 8DOF model. Since the EKF does not need to solve computationally-complex optimization problems (as the MPC does), this is a feasible solution. Additionally, IMU integration may offer further improvements to the state estimation.

We plan to continue MOCAP testing, ensuring that similar noise levels in simulation and reality are implemented. Additionally, we hope to provide tests integrating real GPS measurements, allowing for validation of the GPS noise model described in Sec.~\ref{sec:noiseModel}. The current challenge with regard to this is the inability to accurately generate ground truth data in real world environments. A potential solution we are exploring involves the usage of an RTK GPS, which can improve GPS to have centimeter level accuracy. This would allow for long-distance waypoint following tests, which would better fit the intentions of an EKF state estimator.